%% file: acl_latex.tex
\pdfoutput=1

\documentclass[11pt]{article}

\usepackage{acl}

\usepackage{times}
\usepackage{latexsym}

\usepackage[T1]{fontenc}

\usepackage[utf8]{inputenc}

\usepackage{microtype}

\usepackage{inconsolata}

\usepackage{mathtools}
\mathtoolsset{showonlyrefs,showmanualtags}

\usepackage{tabularx}
\usepackage{multirow}
\usepackage{graphicx}
\usepackage{booktabs}
\usepackage{amsmath}
\usepackage{xcolor}
\usepackage{soul}

\usepackage{pgfplots}
\pgfplotsset{compat=1.18}
\usetikzlibrary{patterns}
\usepackage{tikz}
\usetikzlibrary{positioning}

\renewenvironment{quote}{%
  \list{}{%
    \leftmargin0.1cm   %
    \rightmargin\leftmargin
  }
  \item\relax
}
{\endlist}

\newenvironment{tightquote}
{\vspace{-0.4\baselineskip}\quote}
{\endquote\vspace{-0.4\baselineskip}}

\newenvironment{itemizesquish}[2]{\begin{list}{\labelitemi}{\setlength{\itemsep}{#1}\setlength{\labelwidth}{#2}\setlength{\leftmargin}{\labelwidth}\addtolength{\leftmargin}{\labelsep}}}{\end{list}}

\definecolor{lightblue}{RGB}{173,216,230}
\definecolor{lightpink}{RGB}{229,218,218}
\definecolor{lightyellow}{RGB}{255,241,153}%

\newcommand{\hlyellow}[1]{\sethlcolor{lightyellow}\hl{#1}}
\sethlcolor{yellow}

\title{Can Large Language Models Follow Concept Annotation Guidelines?\\A Case Study on Scientific and Financial Domains}

\author{Marcio Fonseca $\qquad$ Shay B. Cohen \\
Institute for Language, Cognition and Computation \\
School of Informatics, University of Edinburgh \\
10 Crichton Street, Edinburgh, EH8 9AB \\
\medskip
\texttt{m.fonseca@ed.ac.uk},
\texttt{scohen@inf.ed.ac.uk}}

\begin{document}
\maketitle
\begin{abstract}
Although large language models (LLMs) exhibit remarkable capacity to leverage in-context demonstrations, it is still unclear to what extent they can learn new facts or \emph{concept definitions} via prompts. To address this question, we examine the capacity of instruction-tuned LLMs to follow in-context \emph{concept annotation guidelines} for \emph{zero-shot} sentence labeling tasks. We design guidelines that present different types of \emph{factual} and \emph{counterfactual} concept definitions, which are used as prompts for zero-shot sentence classification tasks. Our results show that although concept definitions consistently help in task performance, only the larger models (with 70B parameters or more) have limited ability to work under counterfactual contexts. Importantly, only proprietary models such as \textsc{GPT-3.5} can recognize nonsensical guidelines, which we hypothesize is due to more sophisticated alignment methods. Finally, we find that \textsc{Falcon-180B-chat} is outperformed by \textsc{llama-2-70B-chat} is most cases, which indicates that increasing model scale does not guarantee better adherence to guidelines. Altogether, our simple evaluation method reveals significant gaps in concept understanding between the most capable open-source language models and the leading proprietary APIs.\footnote{Code and dataset are available at \url{https://github.com/thefonseca/concept-guidelines}}
\end{abstract}

\section{Introduction}

Large language models (LLMs) are known to distill knowledge from vast datasets during the pre-training phase \citep{brown2020language}. Such knowledge can be queried via prompting, which allows the application of LLMs to several knowledge-intensive tasks in zero-shot and few-shot settings. In particular, recent work demonstrates promising applications of LLMs to reduce the cost of data annotation in several domains \citep{wang2021want,agrawal2022large,zhu2023can}.

\begin{figure}[t!]
    \centering
\noindent\fbox{%
\parbox{.98\columnwidth}{%
\begingroup
\begin{tightquote}
\hlyellow{\texttt{Consider the following concepts:}}\\
\hlyellow{\texttt{\-- \textbf{Background: A sentence that provides context, foundational knowledge, or relevant information about the research topic, existing theories, prior studies, or the broader scientific field in which the research is situated.}}}\\
\hlyellow{\texttt{\;\;\;\;(more definitions $c_K$: $\delta(c_K)$\ldots)}}\\
\hlyellow{\texttt{\-- \textbf{Conclusion: A sentence that summarizes the key takeaways, implications, interpretations, or insights derived from the study's results.}}}\\
\color{black}
\\
\ul{\texttt{Classify the text below into one of the categories listed above. Be concise and write only the category name.}}\\
\\
\ul{\texttt{Text: Therefore, the phase transition can be classified as essentially driven by Coulomb interactions.}}\\
\ul{\texttt{Concept:}} \texttt{\textbf{Conclusion}}
\end{tightquote}
\endgroup
}}
\caption{An abridged example of zero-shot sentence classification using a \hlyellow{\texttt{concept guideline prompt}}. We perform controlled interventions in \hlyellow{\texttt{\textbf{concept definitions}}} (pairs of concept labels $c_K$ and their descriptions $\delta(c_K)$) while keeping the \texttt{\ul{task prompt}} fixed. We aim to gauge the capacity of the model to learn new concepts during inference, \emph{without in-context demonstrations}.}\label{fig:guideline_prompt}
\end{figure}

Most data labeling efforts based on LLMs leverage in-context demonstrations to elicit the desired concepts. However, previous work suggests that 
language models cannot learn from in-context ground-truth labels but just leverage demonstrations to infer the task format and label space \citep{min2022rethinking}. In contrast, human annotators typically follow \emph{guidelines}, which in addition to examples, include concept definitions \citep{liakata2008guidelines} that complement and modify the annotator's prior concept understanding to align with the labeling goals.

In this work, we assess the capacity of LLMs to follow analogous in-context \emph{concept annotation guidelines} for sentence classification tasks. Our goal is to verify if language models can learn from in-context definitions and change their behavior consistently in downstream tasks \citep{onoe2023can}. To this end, we design several types of guidelines that represent both \emph{factual} and \emph{counterfactual} concept definitions\footnote{In this work, we denote as \emph{counterfactual} those concept definitions that disagree with commonsense understanding, which we assume to be prevalent in a \emph{default world model} \citep{wu2023reasoning} derived from the pre-training data. We formalize \emph{factual} and \emph{counterfactual} guidelines in Section \ref{sec:concept_guidelines}.}. Our assumption is that learning from concept definitions would imply the capacity to reason in contexts that contradict the model's prior knowledge.

In our experiments, we evaluate the \textsc{llama-2} model by \citet{touvron2023llama} (7B, 13B, and 70B-parameter chat variants), \textsc{Falcon-180B-chat} \citep{falcon}, \textsc{GPT-3.5}, and \textsc{GPT-4} \citep{openai2023gpt4} on zero-shot sentence classification tasks (as illustrated in Figure \ref{fig:guideline_prompt}). The tasks require the recognition of scientific concepts, for which labels are likely present in the models' pre-training data. \citep{liakata2008guidelines}. To control for pattern memorization, we also annotate a novel dataset of company disclosures with financial concepts based on the Integrated Reporting framework \citep{cheng2014international}.

In both domains, we observe a consistent classification performance improvement when models have access to concept labels paired with their factual concept definitions (compared to just a list of labels). However, when presented with counterfactual guidelines, only larger models (70B parameters or more) tend to output predictions consistent with guidelines. Still, we observe that scaling alone is not sufficient, as \textsc{Falcon-180B-chat} is outperformed by \textsc{llama-2-70B-chat} in most settings. Importantly, only proprietary models are able to recognize unsolvable tasks, that is, ones for which the guidelines provide nonsensical concept labels. Finally, we find that the performance of more capable models such as \textsc{GPT-3.5} is more strongly correlated to the degree of guideline factuality compared to \textsc{llama-2-7B}, suggesting that the former model has a more nuanced concept understanding.

Overall, some of our findings reinforce previous studies focusing on few-shot learning using perturbed labels \citep{wei2023larger} and chain-of-thought reasoning \citep{saparov2022language}. However, our classification tasks require the model to generalize \emph{only from concept definitions}, without demonstrations. Additionally, unlike previous work, we provide extensive experiments using state-of-the-art open-source models. Although these models may approach the aggregate performance of proprietary APIs, our results reveal important gaps in terms of concept understanding, especially in counterfactual scenarios and regarding the ability to recognize nonsensical tasks. 

\section{Concept Classification with Guidelines}
Let $S$ and $C$ be random variables representing sentences (from the set of token sequences $\mathcal{S}$) and corresponding latent concepts to be inferred (from the concept set $\mathcal{C}$; e.g., whether the sentence conveys scientific \emph{background} or \emph{methods}). To specify the task, we introduce annotation guidelines $G$, which specify concept labels and their definitions. Then, the concept annotation process for a sentence $s$ given the guideline $g$ is formalized as follows:
\begin{align}\label{eq:annotation_process}
    c_s &= \arg\max_{c^\prime \in \mathcal{C}} P(C=c^\prime \mid K, G=g, S=s),
\end{align}
where $c_s$ is the inferred concept and $K$ represents prior domain knowledge about the concepts of interest. Then, we define a language model $P_\theta$ that approximates Eq.~\refeq{eq:annotation_process} through conditional generation:
\begin{align}\label{eq:conditional_generation}
    y_s &= P_\theta(\cdot \mid \texttt{prompt}_G(g); \texttt{prompt}_T(s)),
\end{align}
where $y_s$ is a concept label, that is, a sequence of tokens corresponding to a concept $c_s$. The functions $\texttt{prompt}_G$ and $\texttt{prompt}_T$ are textual templates that describe concept guidelines and a concept classification task respectively (see Figure~\ref{fig:guideline_prompt}). These guidelines and task prompts concatenated condition the language model generation.

The language model parameters $\theta$ capture the prior conceptual knowledge $K$ acquired during pre-training and instruction-tuning. In addition, we hypothesize that $\theta$ encodes the conditional dependency between the guidelines $G$ and the concepts $C$ as the guidelines express relationships between concept labels and their task-specific definitions. By performing controlled interventions in the guidelines, we aim to measure how concept understanding is affected by the following factors: 1) the lexical information of concept labels and concept definitions; 2) the degree of factuality of concept definitions. In the next section, we present guidelines designed to capture such factors.

\subsection{Concept Guidelines}
\label{sec:concept_guidelines}
We associate each concept $c_i \in \mathcal{C}$ to a natural language definition $d_j \in \mathcal{D}$ through a injective \emph{concept definition} function $\delta \colon \mathcal{C} \rightarrow \mathcal{D}$, where each $c_i \in \mathcal{C}$ and $d_j \in \mathcal{D}$ are sequences of tokens. Each association $\delta(c_i) = d_j$ represents a \emph{factual definition} if $i = j$ and a \emph{counterfactual definition} otherwise. Then, a concept guideline is formalized as a tuple $G = \langle \mathcal{C}, \mathcal{D}, \delta\rangle$. Depending on the choice of the function $\delta$, we derive different types of factual and counterfactual concept guidelines, which we describe below.

\paragraph{Factual guidelines $G_f$}
The factual guidelines combine the concept labels with their corresponding factual definitions, that is, $\delta(c_i) = d_i$ for all $i$. To illustrate, a factual guideline prompt would include the following definition for the scientific concept \textsc{background}:

\begin{quote}
    \emph{Background: A sentence that provides context, foundational knowledge, or relevant information about the research topic, existing theories, prior studies, or the broader scientific field in which the research is situated.}
\end{quote}

The factual guideline serves as a control baseline to compare against other types of guidelines. For each of the scientific and financial concepts explored in this paper, we use definitions generated by \textsc{GPT-3.5}. These definitions are further reviewed for quality and redacted to remove explicit mentions of label names. We detail the generation of concept definitions in Section \ref{sec:concept_definitions}.

\paragraph{Out-of-dictionary guidelines $G_{\text{OOD}}$}
We replace real concept labels $c_i$ from factual guidelines with out-of-dictionary (OOD) words such as \texttt{Snizzlewump} and \texttt{Wobblequark}. With those OOD words, we remove the dependency with respect to prior knowledge tied to the lexical information of concept labels. The OOD labels are generated by \textsc{GPT-3.5} using the prompt: \textit{Generate a list of random out-of-dictionary words.} The resulting words are: \texttt{Flibberknock},
\texttt{Quibblesnatch}, \texttt{Blibberflop}, \texttt{Ziggledorf}, \texttt{Snizzlewump}, \texttt{Wobblequark}, \texttt{Jibberplunk}, \texttt{Crumblefluff}, \texttt{Splonglewort},
\texttt{Dinglewhack}.

\paragraph{Empty-definition guidelines $G_\varepsilon$}
As a variant of the factual and OOD guidelines above, we replace each definition with an empty string, that is, $\delta(c_i) = \varepsilon$ for all $i$. We denote these guidelines $G_{f,\varepsilon}$ and $G_{\text{OOD},\varepsilon}$ respectively, and use them to gauge the contribution of concept definitions compared to the factual guideline $G_f$ baseline.

\paragraph{Counterfactual guidelines $G_c$}
A guideline is considered counterfactual when at least one concept $c_i$ is paired with a definition from another concept $c_j$, that is, $\delta(c_i) = d_j$ for $i \neq j$. Since $\delta$ is injective, we have the number of counterfactual definitions (the \emph{degree of counterfactuality} of a guideline) ranging from two to $|\mathcal{C}|$.

\subsection{Concept Definitions}
\label{sec:concept_definitions}

Ideally, we would use the same guidelines provided to humans (e.g., the financial guidelines in Appendix  \ref{sec:financial_annotation_details}) for annotation with LLMs. However, the human guidelines are not uniform across concepts and contain several examples and explicit references to external content. Thus, to minimize the confounding factors related to differences in definitions across concepts in both financial and scientific domains, we use model-generated concept definitions\footnote{For completeness, we also provide experimental results with human guidelines in Appendix \ref{sec:results_human_guidelines}.}. Specifically, we prompt \textsc{GPT-3.5} (refer to Section \ref{sec:models} for API usage details) to provide a short description of a concept in the context of a sentence annotation task. For scientific concepts, we use the following prompt:
\noindent\fbox{%
    \parbox{0.975\columnwidth}{%
\begingroup
\addtolength\leftmargin{-0.1in}

\begin{tightquote}
\texttt{We need to classify sentences in scientific articles according to the information they convey: background, motivation, method, results, or conclusion. Please provide a short definition for each of those labels to be used in annotation guidelines.}
\end{tightquote}
\endgroup
}}
\noindent Then, we review and edit the definitions to remove explicit mentions of label names such as \emph{A sentence is classified as "Motivation" when it explains (...)}. The final scientific concept definitions are provided in Appendix \ref{sec:full_concept_definitions}, Table \ref{tab:scientific_concepts}. Similarly, we generate definitions for financial concepts using the prompt:
\noindent\fbox{%
    \parbox{0.975\columnwidth}{%
\begingroup
\addtolength\leftmargin{-0.1in}

\begin{tightquote}
\texttt{We need to classify sentences in company disclosure reports according to the capital information they convey: financial, manufactured, intellectual, human, social and relationship, or natural. Based on the Integrated Reporting framework, please provide a short definition for each of those labels to be used in annotation guidelines.}
\end{tightquote}
\endgroup
}}
\noindent The financial concept definitions (after review) are provided in Appendix \ref{sec:full_concept_definitions}, Table \ref{tab:financial_concepts}.

\section{Experimental Setup}
\label{section:experiments}
To experiment with different types of guidelines defined in Section \ref{sec:concept_guidelines}, we choose concepts $\mathcal{C}$ for which the language models have exposure via pre-training data. The first domain we explore relates to rhetorical roles in scientific articles (Section \ref{sec:scienfic_concepts}), which is extensively covered in the literature \citep{liakata2012automatic}. Since the pre-training data for LLMs likely include various scientific concept classification datasets, we annotate a novel dataset of sentence-level financial concepts (Section \ref{sec:financial_concepts}). In addition to controlling for label memorization, our financial annotation based on the Integrated Reporting framework \citep{cheng2014international} covers concepts that are technical but arguably more accessible than scientific rhetoric. In Sections \ref{sec:models} and \ref{sec:classfication_details}, we detail the LLM baselines used in the experiments and the classification task hyperparameters.

\subsection{Scientific Concepts Dataset}
\label{sec:scienfic_concepts}
To test a model's knowledge of scientific concepts we use the ARTCorpus dataset \citep{liakata2008guidelines}, which consists of 35,040 sentences from 225 chemistry papers annotated by experts. Each sentence is annotated with one of the 11 \emph{Core Scientific Concepts} (CoreSCs) derived from the EXPO ontology \citep{soldatova2006ontology}. 

In the CoreSC scheme, the scientific concepts are structured hierarchically, with concepts such as \emph{hypothesis}, \emph{motivation}, and \emph{goal} being different sub-types of scientific \emph{objectives}. Since the dataset is relatively small, we observed that some classes were too fine-grained, resulting in a strong label imbalance. To address this issue, we merged some of the categories that shared the same parent concept, yielding the following set of categories: \textsc{Background}, \textsc{Objective}, \textsc{Methods}, \textsc{Results}, and \textsc{Conclusion}. This classification scheme is also used in other PubMed-derived datasets such as the PubMed RCT \citep{dernoncourt2017pubmed}. In our experiments, we use 500 sentences (100 samples per scientific concept) sampled from the ARTCorpus training split.

\subsection{Financial Concepts Dataset}
\label{sec:financial_concepts}
In this section, we introduce the methodology used to collect and annotate a dataset of company disclosures with financial concepts.

\subsubsection{Data Collection}
We collected narrative sections from 10-K annual reports extracted from the Electronic Data Gathering, Analysis, and Retrieval (EDGAR) system \citep{SEC2014}, which is used by companies to submit documents to the United States Securities and Exchange Commission (SEC). For each report, we use the following sections: Item 1 - Business, Item 7 - Management's Discussion and Analysis, and Item 7A - Quantitative and Qualitative Disclosure about Market Risk. The reports are published in December 2021 by companies in the S\&P 500 index \citep{SP500} with the largest market capitalization across 11 industry sectors.

\subsubsection{Annotation Scheme}
Several reporting standards (IFRS\footnote{\url{https://www.ifrs.org}}, GAAP\footnote{\url{https://www.investopedia.com/terms/g/gaap.asp}}) and ontologies such as FIBO \citep{bennett2013financial} have been developed, but they are often too technical and complicated to derive a simple taxonomy of financial concepts. Fortunately, the Integrated Reporting <IR> framework\footnote{\url{https://integratedreporting.org}} offers a suitable set of domain concepts for the task. It defines a set of reporting elements that deliver a holistic view of how the company uses capital to generate value (in this case, value in a broad sense, not just financial). 

In this work, we use one dimension of the <IR> framework related to \emph{capitals}, which is the pool of funds available to an organization for use in the production of goods or the provision of services. The capital concept types are: \textsc{financial}, \textsc{manufactured}, \textsc{intellectual}, \textsc{human}, \textsc{Social and relationship}, and \textsc{natural}.

\subsubsection{Annotation Process}
\label{sec:annotation_process}
In this section, we describe the main steps of the annotation workflow, which include annotator selection and training, an agreement assessment phase, and the final annotation phase. Our annotation process is inspired by the General Scientific Concepts guidelines \citep{liakata2008guidelines}, but our concepts typology is not hierarchical and does not account for instances of concepts (i.e., assigning identifiers for each concept instance).

\paragraph{Hiring and training} Two final-year undergraduate students and one graduate student with a background in finance/economics were hired as annotators. The compensation was 10 British Pounds per hour of work, with an estimated effort of 50 sentences per hour. Each annotator received one-to-one training about the motivation, annotation scheme, and guidelines, which included a description of each financial concept, examples, and general instructions covering edge cases. The full guideline content is provided in Appendix \ref{sec:financial_annotation_details}.

\paragraph{Agreement assessment} In the first round of annotation, each labeler worked on the same report (with 1,291 sentences) and then the agreement was estimated using the weighted Cohen’s $\kappa$ statistic \citep{artstein-poesio-2008-survey}. The weighted version was adopted because each annotator is allowed up to two choices for capitals, so partial agreements are also taken into account. Formally, we define the disagreement weight $d$ as the symmetric difference between the sets of labels $L_{a,i}$ and $L_{b,i}$ assigned to sample $i$ by annotators $a$ and $b$ respectively:
\begin{align*}
d_i(a, b) = |(L_{a,i} \cup L_{b,i}) \setminus (L_{a,i} \cap L_{b,i})|.
\end{align*}
The disagreements $d_i(a,b)$ are used in the weighted Cohen's $\kappa$ formulation by \citet{artstein-poesio-2008-survey}, Section 2.6.2.

\begin{table}
  \centering
  \setlength\tabcolsep{8.8pt}
  \begin{tabular}{l|ccc}
    \toprule
    \multicolumn{1}{c|}{\multirow{2}{2cm}{\centering \textbf{Annotation Round}}} & \multicolumn{3}{c}{\textbf{Annotator Agreement}} \\
    \cmidrule(r){2-4}
     & $A_{12}$ & $A_{13}$ & $A_{23}$ \\
    \toprule
    Round 1 & 0.27 & 0.35 & 0.35 \\
    Round 2 & 0.45 & 0.60 & 0.35 \\
    \bottomrule
  \end{tabular}
  \caption{Annotator agreement on capital labels. $A_{ij}$ is the weighted Cohen's $\kappa$ between annotators $i$ and $j$.}\label{tab:agreement}
\end{table}

The first round aimed at gauging the annotator's understanding of the guidelines and also, collecting feedback to improve the instructions. After analysis of the results, a one-to-one review session was delivered to give feedback about some common misconceptions and disagreements. A second report (562 sentences) was released to assess the effect of the improved guidelines. As shown in Table \ref{tab:agreement}, the scores improved significantly in the second round, suggesting the changes in the guidelines were effective\footnote{Due to the inherent ambiguity of the task, the agreement scores are moderate. We discuss this issue in the limitations section.}. The final concept labels for the first two reports were chosen by majority voting.\footnote{Voting ties were adjudicated by the guideline's author.}

\paragraph{Final annotation} Following \citet{liakata2008guidelines}, the first two phases of annotation are used for quality assessment, and each subsequent report is annotated by just one annotator. Table \ref{tab:annotation_statistics} details the annotation statistics. In our experiments, we use a balanced sample of 540 sentences, with 90 sentences for each of the 6 financial concepts.

\begin{table}
  \centering
  \setlength\tabcolsep{2.4pt}
  \begin{tabular}{l|ccc}
    \toprule
    \textbf{Company} & \textbf{Sector} & \textbf{Sentences} & \textbf{Labelers} \\
    \toprule
    \multicolumn{1}{l|}{\multirow{2}{1.4cm}{Monster Beverage}} & \multicolumn{1}{c}{\multirow{2}{1.4cm}{\centering Consumer Staples}} & \multicolumn{1}{c}{\multirow{2}{1.4cm}{\centering 1291}} & \multicolumn{1}{c}{\multirow{2}{1.4cm}{\centering 3}} \\
    \\
    \midrule
    Chevron & Energy & 562 & 3 \\
    \midrule
    \multicolumn{1}{l|}{\multirow{2}{1.4cm}{Netflix}} & \multicolumn{1}{c}{\multirow{2}{2.3cm}{\centering Communication Services}} & \multicolumn{1}{c}{\multirow{2}{1.4cm}{\centering 648}} & \multicolumn{1}{c}{\multirow{2}{1.4cm}{\centering 1}} \\\\
    \midrule
    \multicolumn{1}{l|}{\multirow{2}{1.4cm}{Amazon}} & \multicolumn{1}{c}{\multirow{2}{2.1cm}{\centering Consumer Discretionary}} & \multicolumn{1}{c}{\multirow{2}{1.4cm}{\centering 609}} & \multicolumn{1}{c}{\multirow{2}{1.4cm}{\centering 1}} \\\\
    \midrule
    \multicolumn{1}{l|}{\multirow{2}{1.4cm}{Sherwin- Williams}} & \multicolumn{1}{c}{\multirow{2}{1.4cm}{\centering Materials}} & \multicolumn{1}{c}{\multirow{2}{1.4cm}{\centering 687}} & \multicolumn{1}{c}{\multirow{2}{1.4cm}{\centering 1}} \\\\
    \bottomrule
  \end{tabular}
  \caption{Statistics for annual reports (published in December 2021) annotated with financial concept labels.}\label{tab:annotation_statistics}
\end{table}

\subsection{Models}
\label{sec:models}
In our experiments, we use the leading open-source and proprietary instruction-tuned language models currently available, covering a wide range of sizes (from 7B to 180B for open-source models). We focus on instruction-tuned models as non-instruct models require the task specification via in-context samples \citep{xie2021explanation,min2022rethinking}, which in our early experiments resulted in poor performance when mixed with concept guidelines.

\begin{itemizesquish}{-0.3em}{0.5em}
    \item \textbf{\textsc{Llama-2}} \citep{touvron2023llama}, a family of open-source large language models that achieve state-of-the-art results at the moment of this writing. We use the \textsc{Llama-2-chat} variants (7B, 13B, and 70B parameters), which are pre-trained on 2 trillion tokens of data and fine-tuned via supervised fine-tuning and Reinforcement Learning with Human Feedback (RLHF). Unless otherwise stated, all mentions of \textsc{Llama-2} in this work refer to the chat variants.
    \item \textbf{\textsc{GPT-3.5}} and \textbf{\textsc{GPT-4}} \citep{openai2023gpt4}, two proprietary models that offer the best instruction-following capabilities at the time of this writing. In our experiments, \textsc{GPT-3.5} and \textsc{GPT-4} refer to the \texttt{gpt-3.5-turbo-0613} and \texttt{gpt-4-0613} models respectively\footnote{\url{https://platform.openai.com/docs/models}}, which are invoked via the chat completions API\footnote{\url{https://platform.openai.com/docs/guides/gpt}}. 
    \item \textbf{\textsc{Falcon-180B}} \citep{falcon}, a 180-billion language model trained on 3.5 trillion tokens from the RefinedWeb dataset \citep{penedo2023refinedweb}. We use the \textsc{falcon-180b-chat} version that is fine-tuned on further instruction, question answering, and chat datasets. In contrast to the other language models above, it does not use Reinforcement Learning with Human Feedback (RLHF) in its fine-tuning phase.
\end{itemizesquish}

\subsection{Concept Classification Details}
\label{sec:classfication_details}
For concept classification, we perform conditional generation (Eq. \refeq{eq:conditional_generation}) using the Hugging Face transformers library \citep{wolf-etal-2020-transformers}. To build the model inputs, we apply the prompt template in Appendix~\ref{sec:inference_details}, replacing the placeholders with the concept labels, definitions, the input sentence, and a concept domain indicator. Since the input sentences are short (around 30 words on average), no truncation is applied. We provide details on the prompts and inference parameters in Appendix \ref{sec:inference_details}. 

\paragraph{Post-processing}
Since we use unconstrained generation for classification, in some instances the output includes extra dialog verbiage and even explanations for the predictions. By examining these outputs, we can gain more detailed insights into the model behavior, for instance, when it refuses to classify sentences with out-of-dictionary labels (refer to details in Section \ref{sec:results}). To extract labels from those outputs, we apply a post-processing heuristic that checks if any of the labels is a substring of the output. If there is a single substring that meets this requirement, it is considered as the prediction\footnote{We also examine all predictions manually to check for edge cases in model outputs.}. Finally, for OOD guidelines, we replace the OOD label predictions with the corresponding factual labels, so the performance metrics can be computed with respect to the ground-truth labels.

\section{Results and Discussion}
\label{sec:results}

\begin{figure*}[ht]
\scalebox{1.0}{
\input{images/scientific_guidelines.tikz}
}
\scalebox{1.0}{
\input{images/financial_guidelines.tikz}
}
\caption{Concept classification accuracy for different \textbf{scientific (top)} and \textbf{financial (bottom)} concept guidelines. In this experiment, the counterfactual guideline $G_c$ is a random permutation where \emph{all concept definitions} are counterfactual. \emph{Empty-Def} refers to the empty-definition factual ($G_{f,\varepsilon}$) and out-of-vocabulary guidelines ($G_{OOD,\varepsilon}$). Error bars represent the 95\% confidence interval and the dashed line indicates the random classifier baseline.}\label{fig:concept_guidelines}
\end{figure*}
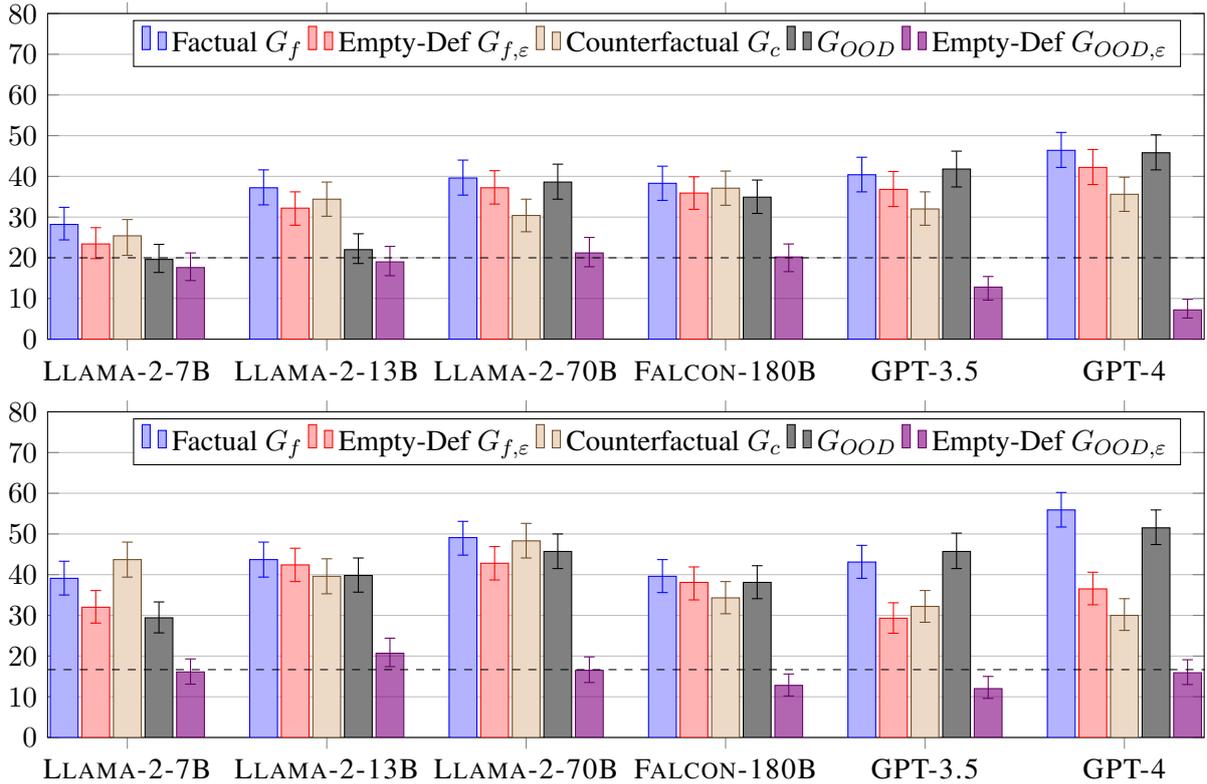

\begin{figure*}[ht]
\scalebox{1.0}{
\input{images/scientific_concept_permutations.tikz}
\input{images/financial_concept_permutations.tikz}
}
\caption{Concept classification accuracy results for different levels of counterfactuality of \textbf{scientific (left)} and \textbf{financial (right)} concept guidelines. We sample 10 guidelines for each counterfactuality level and average the classification accuracies. Error bars represent the standard deviations.}\label{fig:concept_permutations}
\end{figure*}
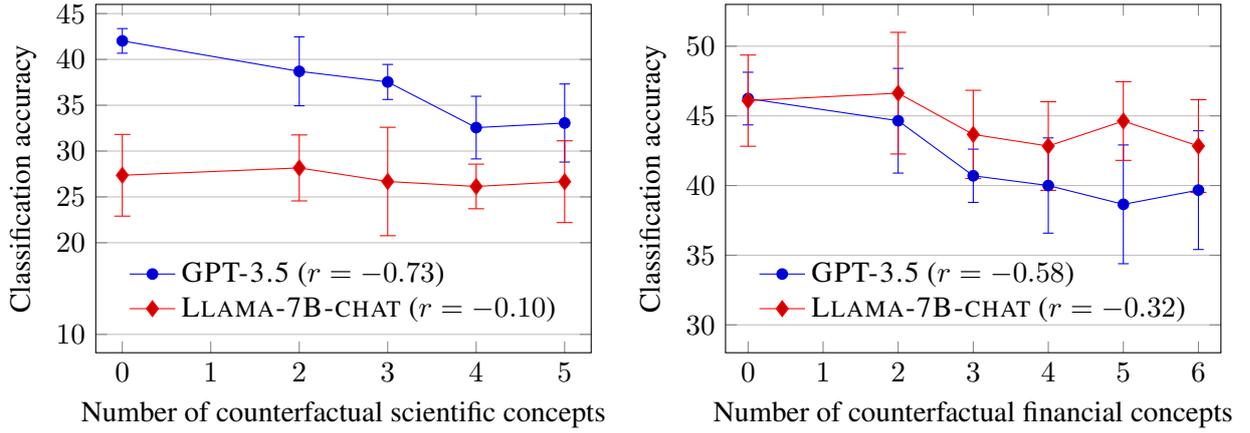

\begin{figure*}[ht]
  \centering
  {\renewcommand{\arraystretch}{0.6}%
  \begin{tabular}{cc}
    \toprule
    \textbf{Financial} & \textbf{Scientific} \\
    \cmidrule(r){1-1} \cmidrule(r){2-2}
     \includegraphics[trim={0.85cm 1.1cm 0cm 0cm},clip,width=0.485\textwidth,height=5.8cm]{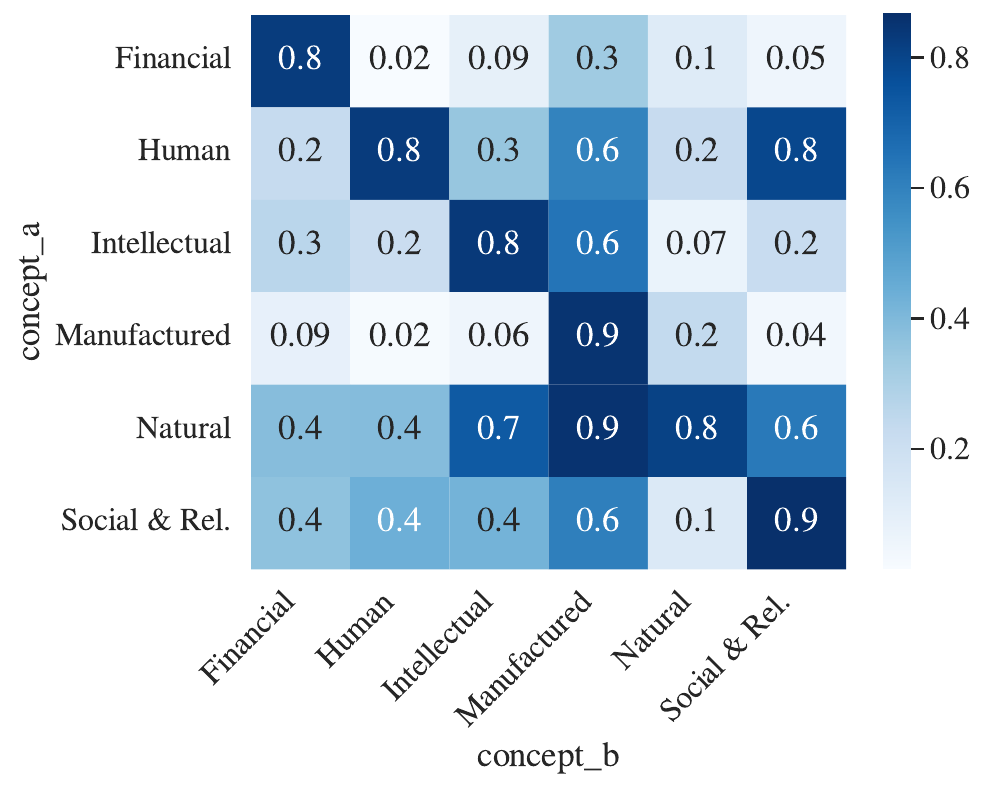} & \includegraphics[trim={0.85cm 1.1cm 0cm 0cm},clip,width=0.485\textwidth,height=5.8cm]{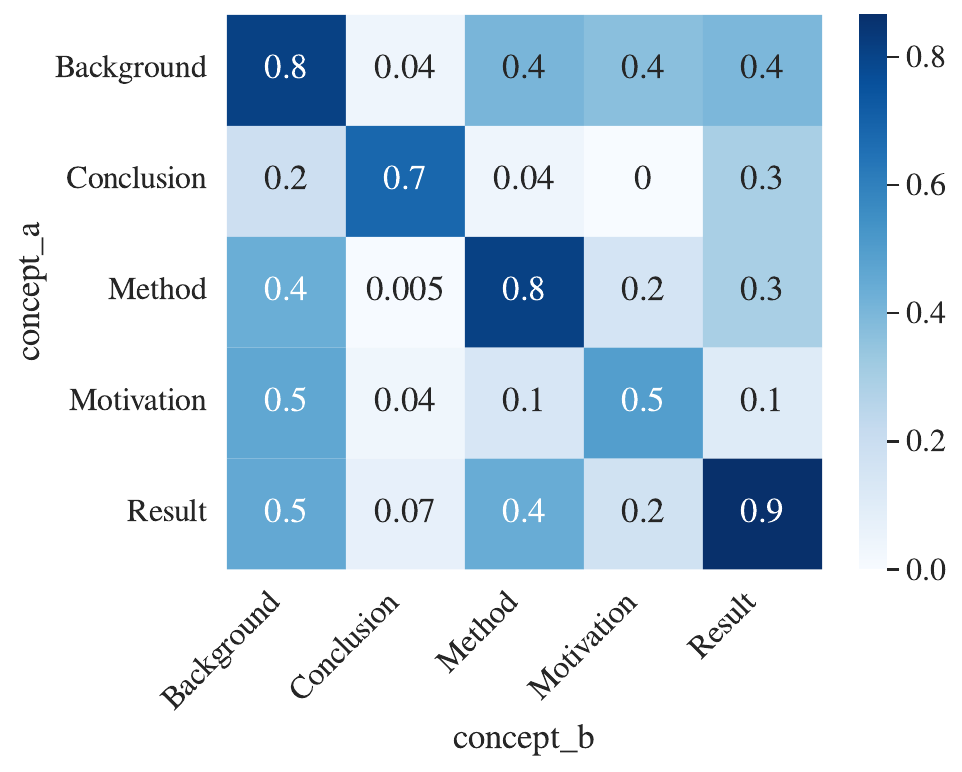} \\
  \end{tabular}
  }
  \caption{Guideline adherence scores per \textbf{financial} and \textbf{scientific} concept for \textbf{\textsc{GPT-3.5}}. Each cell $A_{ij}$ shows the fraction of concept predictions that adhere to concept definitions $\delta(c_j) = d_i$, where the rows indicate original factual labels $c_i$ that are randomly replaced by labels $c_j$ (columns). Off-diagonal results indicate counterfactual definitions.}\label{fig:guideline_effects}
\end{figure*}

\paragraph{LLMs Leverage Concept Labels and Definitions}
Using a factual guideline $G_f$ as a reference, results from Figure \ref{fig:concept_guidelines} show that removing concept definitions (guideline $G_{f,\varepsilon}$) reduces consistently the accuracy of concept classification. However, the classification performance without concept definitions is still significantly higher than the random baseline, which suggests that the models have relevant prior knowledge related to the concept label lexical information. The average accuracy loss when removing concept definitions is $3.7\%$ and $8.2\%$ for scientific and financial concepts respectively, which indicates that financial definitions have a stronger influence on model predictions.

\paragraph{Counterfactual Understanding Emerges with Scaling}
As pointed out above, the lexical information from both concept labels and definitions contributes to task performance. However, we want to verify if the \emph{associations} between labels and definitions are relevant. In Figure \ref{fig:concept_guidelines}, we observe that the smaller \textsc{Llama-2-7B} and \textsc{Llama-2-13B} models have a similar performance under the factual $G_f$ and counterfactual guideline $G_c$ settings. In contrast, there is a consistent drop in accuracy for \textsc{Llama-2-70B}, \textsc{GPT-3.5}, and \textsc{GPT-4}, which indicates that these models are effectively changing the labels according to the counterfactual semantics. Despite having more than two times the number of parameters of \textsc{llama-2-70B}, \textsc{Falcon-180B} behaves similarly to the smaller \textsc{llama-2} models when conditioned with the counterfactual guideline. In this case, scaling is not a sufficient condition to improve understanding in counterfactual contexts.

As further evidence of the capacity of \textsc{GPT} models to follow guidelines, we sample counterfactual concept guidelines such that they are balanced with respect to the number of counterfactual concepts. Then, we evaluate the classification performance for each guideline on the same data samples and average the results for guidelines with the same number of counterfactual concepts. The curves in Figure \ref{fig:concept_permutations} show decreasing classification performance as the scientific guidelines become more counterfactual, with \textsc{GPT-3.5} results having a higher Pearson correlation of $-0.73$ compared to $-0.10$ for \textsc{Llama-2-7B}. A similar trend is observed for financial guidelines, as additional evidence that \textsc{GPT-3.5} is more sensitive to counterfactual guidelines. However, the ability to adhere to counterfactual guidelines is not uniform across concepts. In Figure \ref{fig:guideline_effects}, we observe that some concept changes (e.g., from scientific \textsc{results} to \textsc{conclusion}) are followed much less frequently. We hypothesize that the semantic similarity between concepts may impact the accuracy in counterfactual settings. We leave the study of such factors for future work.

\paragraph{Larger Models Can Rename Existing Concepts}
We consider the effects of removing the lexical information from concept labels by using out-of-dictionary labels ($G_{\text{OOD}}$ guideline). Again, the largest models (70B or more parameters) tend to perform on par with the original factual guideline $G_f$ across both domains. Even though the models are not learning entirely new concepts, it is remarkable that they can associate novel labels with abstract concepts and leverage them to solve tasks. We believe this ability might be relevant for natural language reasoning problems that require symbolic formulation \citep{pan2023logic}.

\paragraph{Proprietary Models Recognize Unknown Concepts}
While \textsc{Llama-2-70B} has performance similar to \textsc{GPT-3.5} on most settings, when presented with out-of-dictionary (OOD) labels without definitions ($G_{\text{OOD}, \varepsilon}$ guideline), it predicts labels randomly. This result confirms that OOD labels provide no information related to scientific or financial concepts. However, \textsc{GPT-3.5} and \textsc{GPT-4} behave differently, often refusing to assign concepts to the sentences and instead generating outputs such as \emph{None of the categories listed above are appropriate for classifying the given text}. For instance, \textsc{GPT-3.5} refuses to classify 58\% and 51\% of sentences from scientific and financial documents respectively, while the open-source models always predict one of the nonsensical labels. We hypothesize that this ability to recognize unknown concepts is derived from careful alignment efforts \citep{ouyang2022training}, which presents an avenue for improving open-source language models.

\paragraph{Agreement with Human Annotators}
Using sample sentences from the second report of the financial annotation, we measure the agreement of the models' financial concept predictions (using factual concept guidelines) to each human annotator. We find that \textsc{Llama-2-7b} and \textsc{GPT-4} achieve average Cohen's $\kappa$ scores on par with expert annotators (Table \ref{tab:model_agreement}). This result is in line with previous work showing that LLMs can be a useful tool in annotation pipelines \citep{wang2021want}.

\begin{table}[ht]
  \centering
  \setlength\tabcolsep{5.8pt}
  \begin{tabular}{l|cccc}
    \toprule
    \multicolumn{1}{c|}{\multirow{2}{2cm}{\centering \textbf{Model}}} & \multicolumn{4}{c}{\textbf{Annotation Agreement}} \\
    \cmidrule(r){2-5}
     & $A_{1}$ & $A_{2}$ & $A_{3}$ & Avg \\
    \toprule
    Human Average & 0.46 & 0.43 & \textbf{0.37} & 0.42 \\
    \midrule
    \textsc{Llama-2-7B} & 0.46 & 0.47 & 0.35 & 0.43 \\
    \textsc{GPT-4} & \textbf{0.47} & \textbf{0.53} & 0.35 & \textbf{0.45} \\
    \bottomrule
  \end{tabular}
  \caption{Financial concept annotation agreement to annotators $A_1$, $A_2$, and $A_3$. Results are non-weighted Cohen's $\kappa$ on a subset of sentences for which human annotators assigned at least one capital concept.}\label{tab:model_agreement}
\end{table}

\section{Related Work}
Previous work examined if LLMs exhibit human-like conceptual \emph{grounding} \citep{piantadosi2022meaning}. \citet{patel2021mapping} demonstrate that LLMs such as \textsc{GPT-3} can generalize spatial and color concepts in some settings. Using several counterfactual reasoning tasks such as arithmetic, chess, and drawing\, \citet{wu2023reasoning} show that some proprietary models have limited capacity for reasoning under counterfactual conditions. Our work explores concept classification tasks that are more abstract than spatial concepts but still simpler than the more complex tasks proposed by \citet{wu2023reasoning}. As a consequence, we can more precisely control the level of counterfactuality of the tasks while keeping the same level of difficulty.

A variety of approaches related to editing factual knowledge of LLMs has been explored recently \citep{onoe2023can,meng2022locating,zhu2020modifying}. This line of work proposes different ways to edit memory related to entities and assess if the model outputs in different contexts are consistent with the newly introduced facts. Those approaches focus on updating model parameters while we examine model behavior under in-context concept edits. \citet{min2022rethinking} study the role of in-context demonstrations for various classification tasks. They conclude that associations of samples and labels do not strongly influence the model's performance, suggesting that non-instruct LLMs cannot learn new information from the demonstrations. In contrast, our results suggest that instruction-tuned models are consistently influenced by in-context concept definitions.

Our evaluation protocol is similar to \citet{wei2023larger} work as they use flipped and ``semantically-unrelated'' labels in task demonstrations. While they focus on in-context learning, our experiments are zero-shot tasks including \emph{only concept definitions}. Thus, our setting is arguably harder, requiring the models to generalize from guidelines (not examples) that are relatively agnostic with respect to the classification task. Furthermore, we put a significant effort into evaluating open-source models that are state-of-the-art at the time of this submission. To our knowledge, this kind of evaluation is not addressed by previous work and is relevant to inform the improvement of open-source initiatives.

The potential of LLMs as zero-shot and few-shot data annotators has been demonstrated in medical \citep{agrawal2022large}, social science \citep{zhu2023can}, and other language understanding tasks \citep{wang2021want}. Our work provide further evidence that instruction-tuned models can perform concept classification with agreement scores comparable to expert annotators. Additionally, we show that similarly to humans, LLMs can leverage concept guidelines to improve the annotation quality.

\section{Conclusion}
By using factual and counterfactual concept guidelines for sentence classification, we demonstrate measurable gaps in concept understanding between leading open-source and proprietary instruction-tuned models. While some level of counterfactual concept understanding emerges with scaling, open-source models cannot recognize nonsensical (out-of-dictionary) guidelines, which the closed APIs can address more consistently. One question to be addressed in future work would be to investigate potential correlations between the capacity of reasoning in counterfactual contexts and other common generation issues such as hallucination.

\section*{Limitations}
\paragraph{Opacity of Proprietary Models} 
The experimental results from Section \ref{sec:results} confirm that the proprietary models excel in almost all classification settings. However, we cannot determine if the main cause for the best performance is the scale, training data, or fine-tuning methods since we do not have access to their implementation details.

\paragraph{Inference costs for Large Models} 
Our experiments are severely limited by the computing requirements of the larger open-source LLM models. For instance, \textsc{Falcon-180B-chat} requires around 400GB of memory for inference, equivalent to 5 Nvidia A100-80GB GPUs. Thus, we limit our counterfactual guideline experiments (Figure \ref{fig:concept_permutations}) to include only a subset of possible permutations for \textsc{llama-2-7b} and \textsc{GPT-3.5}.

\paragraph{Consequences of Counterfactual Performance to Other Tasks}
In this work, we measure the capacity of several language models to work under counterfactual contexts. Future investigation efforts could explore how this ability correlates to a potential reduction of hallucinations in generative tasks or even improved performance in natural language reasoning problems \citep{pan2023logic}.

\paragraph{Financial Annotation Agreement}
Due to the ambiguity of financial annotation the task, the agreement scores we report in Section \ref{sec:annotation_process} are relatively moderate (average $\kappa=0.47$ for the second round in Table \ref{tab:agreement}). One of the main factors for disagreement is that some sentences are complex and may contain multiple capitals, as illustrated in the following example (passages conveying capitals are \underline{underlined}):

\begin{quote}
\emph{Such factors include the duration and scope of the pandemic, including any resurgences of the pandemic, and the impact on our \underline{workforce} and operations; the negative impact of the pandemic on the economy and economic activity, including travel restrictions and prolonged low demand for our products; the ability of our \underline{affiliates, suppliers and partners} to successfully navigate the impacts of the pandemic; the actions taken by governments, businesses and individuals in response to the pandemic; the actions of OPEC and other countries that otherwise impact supply and demand and correspondingly, commodity prices; the extent and duration of recovery of economies and demand for our products after the pandemic subsides; and Chevron’s ability to keep its cost model in line with changing demand for our products.}
\end{quote}

In many cases, annotators chose a non-intersecting subset of the capitals, which counts as a disagreement (even though both are partially correct). Those voting ties were reviewed and adjudicated by the author of the guidelines. Previous scientific annotation projects like \citep{liakata2012automatic} also report a moderate agreement ($\kappa=0.55$, median of the best annotators), which demonstrates the difficulty in annotating technical documents.

Finally, even though report agreements between LLM annotations and humans in Table \ref{tab:model_agreement}, our experiments are not designed to fairly compare annotation quality. Before annotation, humans received training and guidelines that are more comprehensive than LLM guidelines. Secondly, humans were able to ``calibrate'' their labels according to previously annotated sentences, whereas LLMs do not have access to this memory.

\section*{Acknowledgements}
This work was supported by Actelligent Capital and used the Baskerville UK National Tier-2 HPC  (https://www.baskerville.ac.uk) at the University of Birmingham. We also thank the anonymous reviewers for their insightful feedback.

\bibliography{custom}

\appendix

\section{Results with Human Guidelines}
\label{sec:results_human_guidelines}

\begin{figure}[ht]
\input{images/financial_human_guidelines.tikz}
\caption{Concept classification accuracy for different \textbf{financial} concept guidelines, using the \emph{same definitions provided to human labelers} (Figure \ref{fig:financial_annotation_guidelines}). In this experiment, the counterfactual guideline $G_c$ is a random permutation where \emph{all concept definitions} are counterfactual. \emph{Empty-Def} refers to the empty-definition factual ($G_{f,\varepsilon}$) and out-of-vocabulary guidelines ($G_{OOD,\varepsilon}$). Error bars represent the 95\% confidence interval and the dashed line indicates the random classifier baseline.}\label{fig:concept_human_guidelines}
\end{figure}
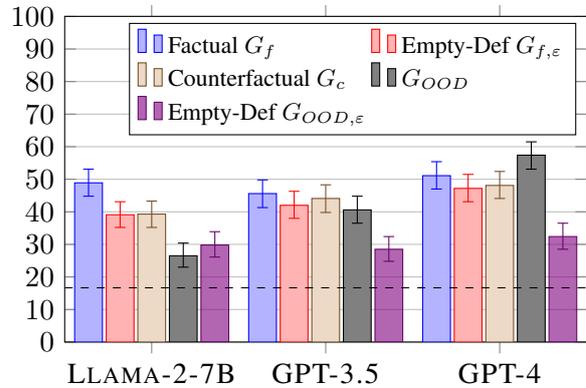

To complement the results in Section \ref{sec:results}, we provide classification accuracies and agreement scores for guidelines using the same definitions provided to human annotators. We observe that models tend to ignore the concept definitions in favor of their prior knowledge about financial concepts, thus reducing the effects of our counterfactual guidelines. This effect is reflected in the more uniform accuracy results shown in Figure~\ref{fig:concept_human_guidelines} (compared to Figure \ref{fig:concept_guidelines}). The human guidelines also result in a slight increase in agreement with human annotators, as shown in Table \ref{tab:model_agreement_human_guidelines} (compared to Table \ref{tab:model_agreement}).

\begin{table}[ht]
  \centering
  \setlength\tabcolsep{5.8pt}
  \begin{tabular}{l|cccc}
    \toprule
    \multicolumn{1}{c|}{\multirow{2}{2cm}{\centering \textbf{Model}}} & \multicolumn{4}{c}{\textbf{Annotation Agreement}} \\
    \cmidrule(r){2-5}
     & $A_{1}$ & $A_{2}$ & $A_{3}$ & Avg \\
    \toprule
    Human Average & 0.46 & 0.43 & \textbf{0.37} & 0.42 \\
    \midrule
    \textsc{Llama-2-7B} & 0.50 & 0.46 & 0.41 & 0.46 \\
    \textsc{GPT-4} & \textbf{0.51} & \textbf{0.57} & 0.42 & \textbf{0.50} \\
    \bottomrule
  \end{tabular}
  \caption{Financial concept annotation agreement to annotators $A_1$, $A_2$, and $A_3$, using the \emph{same definitions provided to human labelers} (Figure \ref{fig:financial_annotation_guidelines}). Results are non-weighted Cohen's $\kappa$ on a subset of sentences for which human annotators assigned at least one capital concept.}\label{tab:model_agreement_human_guidelines}
\end{table}

\section{Concept Definitions}
\label{sec:full_concept_definitions}

The Tables \ref{tab:scientific_concepts} and \ref{tab:financial_concepts} present the definitions for scientific and financial concepts used in our experiments.

\begin{table*}
  \centering
  \begin{tabular}{c|p{0.75\textwidth}}
    \toprule
    \textbf{Concept} & \multicolumn{1}{c}{\textbf{Definition}} \\
    \toprule
    Background & A sentence that provides context, foundational knowledge, or relevant information about the research topic, existing theories, prior studies, or the broader scientific field in which the research is situated. It helps readers understand the background against which the research is conducted. \\
    \midrule
    Motivation & A sentence that explains the reasons, objectives, or goals behind the research. It often includes statements about the research gap, the problem being addressed, the significance of the study, and why the research is important. \\
    \midrule
    Method & A sentence that describes the research methods, techniques, procedures, and data collection processes used in the study. This category also encompasses details about the experimental design, data analysis, and any materials or instruments utilized. \\
    \midrule
    Result & A sentence that presents the empirical findings, outcomes, observations, or data generated by the research. It includes quantitative and qualitative results, statistical analyses, tables, figures, and any other information related to the research findings. \\
    \midrule
    Conclusion & A sentence that summarizes the key takeaways, implications, interpretations, or insights derived from the study's results. It often discusses the broader significance of the findings, suggests future research directions, and may reiterate the study's contributions to the field. \\
    \bottomrule
  \end{tabular}
  \caption{Scientific concept definitions used in sentence classification guidelines.}\label{tab:scientific_concepts}
\end{table*}

\begin{table*}
  \centering
  \begin{tabular}{>{\centering}p{0.125\textwidth}|p{0.74\textwidth}}
    \toprule
    \textbf{Concept} & \multicolumn{1}{c}{\textbf{Definition}} \\
    \toprule
    Financial & A sentence that pertains to monetary resources, assets, liabilities, revenues, expenses, or any other financial information related to the company's operations, investments, and financial performance. \\
    \midrule
    Manufactured & A sentence that refers to physical assets, infrastructure, and tangible resources such as buildings, machinery, equipment, or any other manufactured or constructed items that contribute to the company's value. \\
    \midrule
    Intellectual & A sentence that relates to intangible assets, knowledge, intellectual property, patents, trademarks, copyrights, research and development activities, or any other intellectual assets that enhance the company's competitiveness and innovation. \\
    \midrule
    Human & A sentence that involves information about the company's workforce, including employees, skills, expertise, training, recruitment, talent development, and any other human resources aspects that contribute to the company's success. \\
    \midrule
    Social and \,relationship & A sentence that deals with the company's relationships and interactions with external stakeholders, communities, customers, suppliers, partners, and any other social or relationship-based assets that affect the company's operations and reputation. \\
    \midrule
    Natural & A sentence that addresses environmental resources, sustainability efforts, ecological impacts, conservation initiatives, or any other aspects related to the company's use of natural resources and its environmental responsibility. \\
    \bottomrule
  \end{tabular}
  \caption{Financial concept definitions used in sentence classification guidelines.}\label{tab:financial_concepts}
\end{table*}

\section{Inference Prompts and Parameters}
\label{sec:inference_details}
The prompt illustrated in Figure \ref{fig:guideline_prompt}\footnote{The prompt in Figure \ref{fig:guideline_prompt} is simplified to improve readability.} consists of a guideline prompt $\texttt{prompt}_G$ and a task prompt $\texttt{prompt}_T$. The content of $\texttt{prompt}_G$ is a list of concept definitions:

\noindent\fbox{%
    \parbox{0.975\columnwidth}{%
\begingroup
\addtolength\leftmargin{-0.1in}

\begin{tightquote}
Consider the following concept categories:\\
\texttt{\-- \{$c_1$\}: \{$\delta(c_1)$\}}\\
\textbf{\texttt{\ldots}}\\
\texttt{\-- \{$c_K$\}: \{$\delta(c_K)$\}}
\end{tightquote}
\endgroup
}}
\noindent where $\delta(c_K)$ is a function that maps the concept label $c_K$ to its definition. Then, we define the content of $\texttt{prompt}_T$ as follows:
\noindent\fbox{%
    \parbox{0.975\columnwidth}{%
\begingroup
\addtolength\leftmargin{-0.1in}

\begin{tightquote}
\texttt{Classify the text below into one of the categories listed above. Be concise and write only the category name.}\\
\\
\texttt{Text: \{input sentence $s$\}}\\
\texttt{\{domain\} Concept:}
\end{tightquote}
\endgroup
}}
\noindent where the placeholder \texttt{\{domain\}} is replaced with the text \texttt{Scientific} for scientific concepts and the empty string for financial concepts.

Finally, the prompts above are wrapped into model-specific prompts. For the \textsc{llama-2} models, we use the following prompt:

\noindent\fbox{%
    \parbox{0.975\columnwidth}{%
\begingroup
\addtolength\leftmargin{-0.1in}

\begin{tightquote}
\texttt{[INST] \{$\texttt{prompt}_G$\}}\\
\\
\texttt{\{$\texttt{prompt}_T\}$ [/INST]}
\end{tightquote}
\endgroup
}}
And for \textsc{Falcon-180B}, we use the following prompt:

\noindent\fbox{%
    \parbox{0.975\columnwidth}{%
\begingroup
\addtolength\leftmargin{-0.1in}

\begin{tightquote}
\texttt{User:}
\texttt{\{instruction\} \{$\texttt{prompt}_G$\}}\\
\\
\texttt{\{$\texttt{prompt}_T\}$}\\
\texttt{Falcon:}
\end{tightquote}
\endgroup
}}
Note that we do not use system prompts for both models, as we found that system prompts result in more verbose outputs. The main parameters for inference are provided in Table \ref{tab:generation_parameters}.

\begin{table}
  \centering
  \setlength\tabcolsep{2.9pt}
  \begin{tabular}{lc}
    \toprule
    \multicolumn{2}{c}{\textbf{\textsc{Llama-2}}} \\
    \midrule
    Number of parameters & 7B / 13B / 70B \\
    Max context length & 4096 \\
    \midrule
    \multicolumn{2}{c}{\textbf{\textsc{Falcon-180B-chat}}} \\
    \midrule
    Number of parameters & 180B \\
    Max context length & 2048 \\
    \midrule
    \multicolumn{2}{c}{\textbf{\textsc{Llama-2}} and \textbf{\textsc{Falcon-180B-chat}}} \\
    \midrule
    Parameter type & float16 \\
    Nucleus temperature & 0.8 \\
    Nucleus top-$p$ & 0.95 \\
    \midrule
    \multicolumn{2}{c}{\textbf{\textsc{GPT-3.5}} and \textbf{\textsc{GPT-4}}} \\
    \midrule
    Model \textsc{GPT-3.5} & gpt-3.5-turbo-0613 \\
    Model \textsc{GPT-4} & gpt-4-0613 \\
    temperature & 1 \\
    top\_p & 1 \\
    presence\_penalty & 0 \\
    frequency\_penalty & 0 \\
    \midrule
    \multicolumn{2}{c}{\textbf{All models}} \\
    \midrule
    Max generation tokens & 128 \\
    \bottomrule
  \end{tabular}
  \caption{Summary of generation details and parameters.}\label{tab:generation_parameters}
\end{table}

\section{Guidance Adherence Results}
\label{sec:guideline_effects_llama}
\begin{figure}[ht]
  \centering
  {\renewcommand{\arraystretch}{0.6}%
  \begin{tabular}{c}
    \toprule
    \textbf{Financial} \\
    \includegraphics[trim={0.85cm 0.8cm 0cm 0cm},clip,width=0.485\textwidth,height=5.8cm]{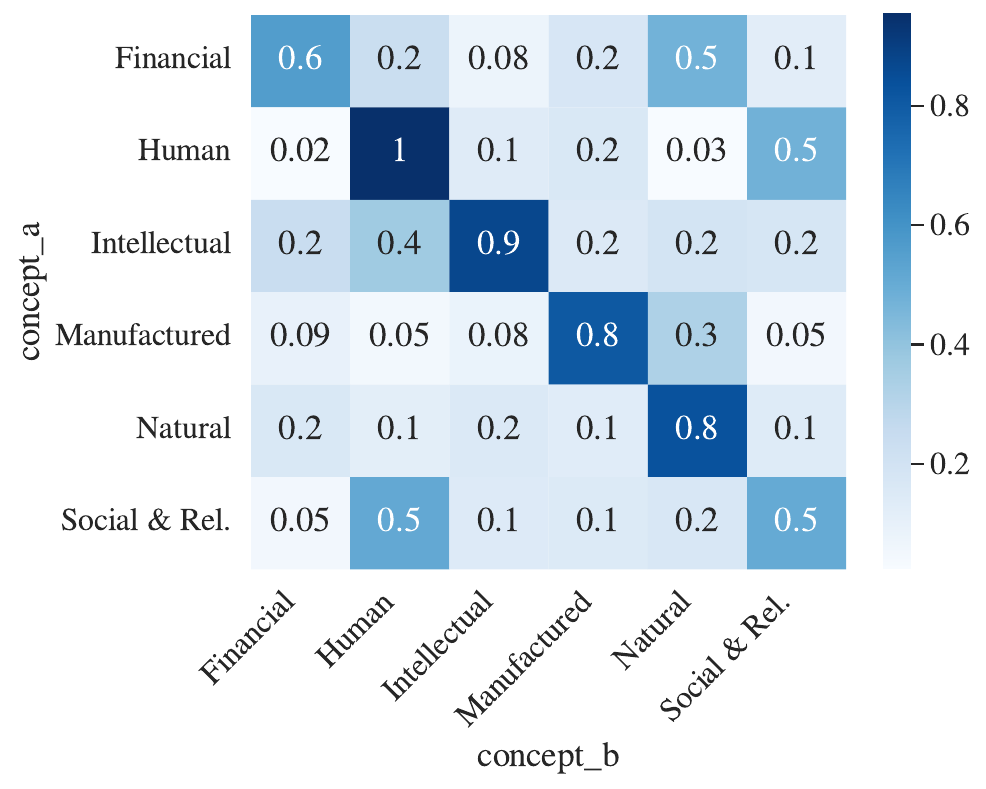} \\
    \midrule
    \textbf{Scientific} \\
    \includegraphics[trim={0.85cm 0.8cm 0cm 0cm},clip,width=0.485\textwidth,height=5.8cm]{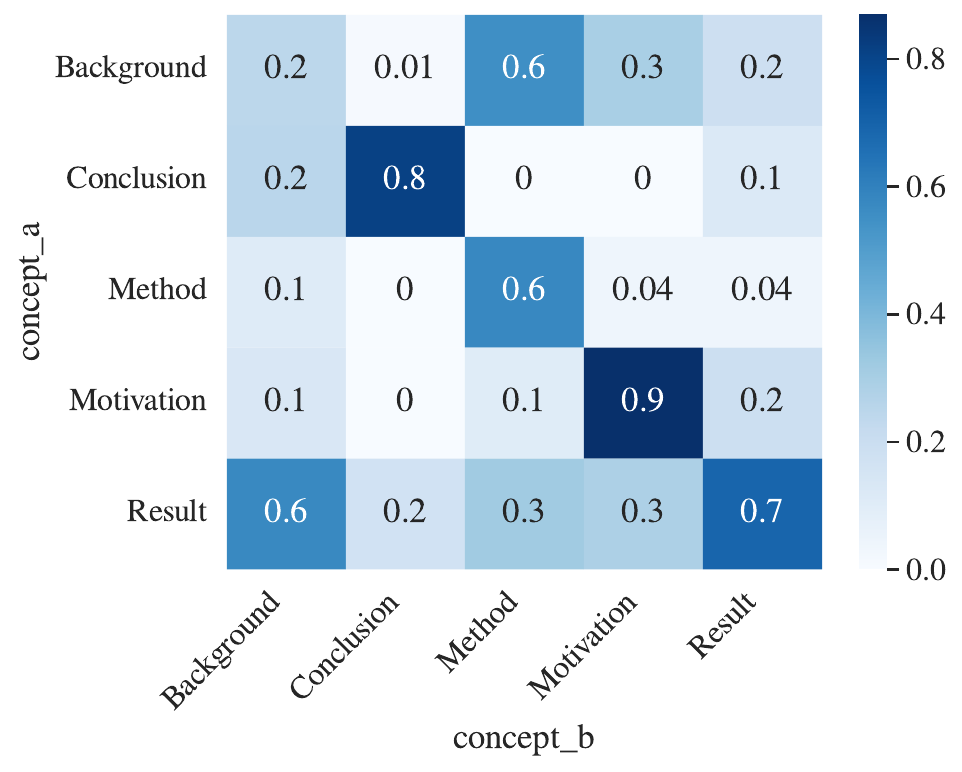} 
     \end{tabular}
  }
  \caption{Guideline adherence scores per \textbf{financial (top)} and \textbf{scientific (bottom)} concept for \textbf{\textsc{llama-2-7b}}. Each cell $A_{ij}$ shows the fraction of concept predictions that adhere to concept definitions $\delta(c_j) = d_i$, where the rows indicate original factual labels $c_i$ that are randomly replaced by labels $c_j$ (columns). Off-diagonal results correspond to counterfactual concept definitions.}\label{fig:guideline_effects_llama}
\end{figure}

To complement the results in Figure \ref{fig:guideline_effects}, we provide the guidance adherence metrics for \textsc{Llama-2-7B} in Figure \ref{fig:guideline_effects_llama}.

\section{Financial Annotation Details}
\label{sec:financial_annotation_details}
\begin{figure*}[ht]
\centering
\includegraphics[trim={0cm 21cm 0cm 0cm},clip,width=0.98\textwidth]{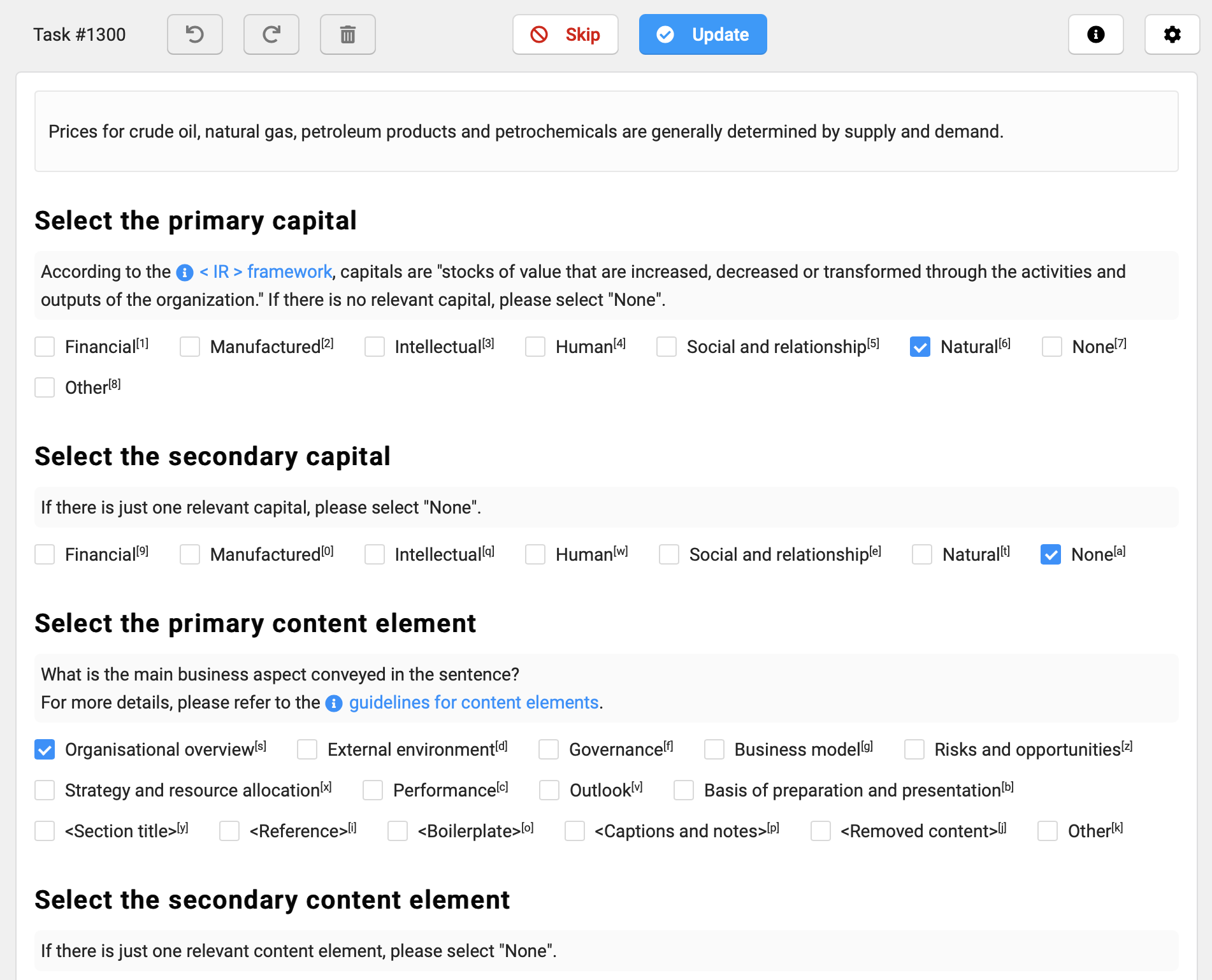}
\caption{The annotation interface for the annotation of financial concepts. Given a sample sentence, annotators are requested to assign one of six capital concepts or \texttt{None}, if not applicable.}
\label{fig:annotation_interface}
\end{figure*}

Before engaging in the annotation task, the hired annotators were presented with the textual guidelines listed in Figure \ref{fig:financial_annotation_guidelines}. The web-based annotation interface (Figure \ref{fig:annotation_interface}) is implemented using Label Studio\footnote{\url{https://github.com/HumanSignal/label-studio}}. The interface shows the sample sentence and requests the annotator to classify it in one of the six capital concepts (\texttt{Financial}, \texttt{Manufactured}, \texttt{Intellectual}, \texttt{Human}, \texttt{Social and relationship}, and \texttt{Natural}) or \texttt{None} if the content is not related to any capital. The annotator also has the option to indicate a secondary capital, if applicable.

The annotation tasks were performed in 2021, when the UK minimum wage was 8.91 British Pounds\footnote{\url{https://www.gov.uk/national-minimum-wage-rates}}, and annotators received 10 British Pounds per hour of work. The experiment design and conditions went through formal approval by an internal ethics committee. The data was not released or stored in public servers to avoid potential contamination.

\begin{figure*}
  \centering
  \begin{tabular}{p{0.97\textwidth}}
    \toprule
    The concepts described in this section follow closely the definitions of the International <IR> framework, and should be sufficient to perform the annotation. According to the IR framework, capitals are “stocks of value that are increased, decreased or transformed through the activities and outputs of the organization.” They can be classified in financial, manufactured, intellectual, social and relationship, and human (<IR> framework, section 2C).\\\\
    
    \textbf{Financial capital}\\
    The pool of funds that is available to an organization for use in the production of goods or the provision of services. It can be obtained through financing or generated through operations and investments. Example: \emph{“The discussion also provides information about the financial results of our business segments to provide a better understanding of how those segments and their results affect the financial condition and results of operations of Ameren as a whole.”}
    \\\\
    \textbf{Manufactured capital}\\ 
    Manufactured physical objects (excluding natural physical objects) that are available to an organization for use in the production of goods or the provision of services, including, buildings, equipment, and infrastructure (such as roads, ports, bridges, etc). Example: \emph{“Due to the long lead time for the manufacture, repair, and installation of the components, the energy center is expected to return to service in late June or early July 2021.”}
    \\\\
    \textbf{Intellectual capital}
    
    Organizational, knowledge-based intangibles, including: Intellectual property, such as patents, copyrights, software, rights and licences “Organizational capital” such as tacit knowledge, systems, procedures and protocols. Example: \emph{“The absence of revenues from a software licensing agreement with Ameren Missouri decreased margins \$5 million.”}
    \\\\
    \textbf{Human capital}
    
    People’s competencies, capabilities and experience, and their motivations to innovate, including their: 1) alignment with and support for an organization’s governance framework, risk management approach, and ethical values; 2) sbility to understand, develop and implement an organization’s strategy; 3) loyalties and motivations for improving processes, goods and services; 4) Other matters related to people management. Example: \emph{“As the situation rapidly evolved, we remained focused on safely serving our customers and protecting the health and safety of our employees.”}
    \\\\
    
    \textbf{Social and relationship capital}
    
    The institutions and the relationships within and between communities, groups of stakeholders and other networks, including: 1) shared norms, and common values and behaviours; 2) key stakeholder relationships, and the trust and willingness to engage that an organization has developed and strives to build and protect with external stakeholders; 3) intangibles associated with the brand and reputation that an organization has developed; 4) an organization’s social licence to operate. Example: \emph{“In March 2020, the MoPSC issued an order in Ameren Missouri’s July 2019 electric service regulatory rate review, approving nonunanimous stipulation and agreements.”}
    \\\\
    
    \textbf{Natural capital}
    
    All renewable and non-renewable environmental resources and processes that provide goods or services that support the past, current or future prosperity of an organization, including air, water, land, minerals, and biodiversity. Example: \emph{“These amounts include the 700 MWs of wind generation projects discussed below, which will support Ameren Missouri’s compliance with the state of Missouri’s requirement of achieving 15\% of native load sales from renewable energy sources beginning in 2021.”}\\
    \bottomrule
  \end{tabular}
  \caption{Guidelines for financial concept annotation provided to human labelers.}\label{fig:financial_annotation_guidelines}
\end{figure*}

\end{document}

%% file: images/scientific_guidelines.tikz
\begin{tikzpicture}
 
\begin{axis} [ybar = .05cm,
    bar width = 10.4pt,
    xmin = 0,
    xmax = 5,
    ymin = 0,
    ymax = 80,
    xtick = data,
    xticklabels = {\textsc{Llama-2-7B},
    \textsc{Llama-2-13B}, \textsc{Llama-2-70B}, \textsc{Falcon-180B}, \textsc{GPT-3.5}, \textsc{GPT-4}},
    ytick distance=10,
    ymajorgrids = true,
    enlarge x limits = {abs = .4},
    enlarge y limits = {abs = .02, upper},
    legend cell align=left,
    legend style={legend columns=5},
    height=5.9cm,
    width=16.8cm,
]
\addplot+ [
    error bars/.cd,
    y dir=both, y explicit,
] coordinates {
(0,28.2) += (0,4.2) -= (0,3.8)
(1,37.2) += (0,4.4) -= (0,4.2)
(2,39.6) += (0,4.4) -= (0,4.2)
(3,38.3) += (0,4.2) -= (0,4.2)
(4,40.4) += (0,4.3) -= (0,4.2)
(5,46.4) += (0,4.4) -= (0,4.2)
};
\addplot+ [
    error bars/.cd,
    y dir=both, y explicit,
] coordinates {
(0,23.4) += (0,4) -= (0,3.6)
(1,32.2) += (0,4) -= (0,4.2)
(2,37.2) += (0,4.2) -= (0,4)
(3,35.9) += (0,4) -= (0,4)
(4,36.8) += (0,4.4) -= (0,4.2)
(5,42.2) += (0,4.4) -= (0,4.2)
};
\addplot+ [
    error bars/.cd,
    y dir=both, y explicit,
] coordinates {
(0,25.4) += (0,4) -= (0,4.8)
(1,34.4) += (0,4.2) -= (0,4.2)
(2,30.4) += (0,4) -= (0,4)
(3,37.1) += (0,4.2) -= (0,4.2)
(4,32) += (0,4.2) -= (0,4)
(5,35.6) += (0,4.2) -= (0,4.2)
};
\addplot+ [
    error bars/.cd,
    y dir=both, y explicit,
] coordinates {
(0,19.6) += (0,3.7) -= (0,3.2)
(1,22) += (0,3.9) -= (0,3.4)
(2,38.6) += (0,4.4) -= (0,4.2)
(3,34.9) += (0,4.2) -= (0,4)
(4,41.8) += (0,4.4) -= (0,4.4)
(5,45.8) += (0,4.4) -= (0,4.2)
};
\addplot+ [
    fill opacity=0.6,
    error bars/.cd,
    y dir=both, y explicit,
] coordinates {
(0,17.6) += (0,3.6) -= (0,3.2)
(1,19) += (0,3.8) -= (0,3.4)
(2,21.2) += (0,3.8) -= (0,3.4)
(3,20.2) += (0,3.2) -= (0,3.6)
(4,12.8) += (0,2.6) -= (0,3.2)
(5,7.2) += (0,2.6) -= (0,2)
};

\legend {Factual $G_f$, Empty-Def $G_{f,\varepsilon}$, Counterfactual $G_c$, $G_{OOD}$, Empty-Def $G_{OOD,\varepsilon}$};

\coordinate (A) at (axis cs:0,20);
\coordinate (O1) at (rel axis cs:0,0);
\coordinate (O2) at (rel axis cs:5,0);
\draw [black,sharp plot,dashed] (A -| O1) -- (A -| O2);

\end{axis}
 
\end{tikzpicture}

%% file: images/financial_guidelines.tikz
\begin{tikzpicture}
 
\begin{axis} [ybar = .05cm,
    bar width = 10.4pt,
    xmin = 0,
    xmax = 5,
    ymin = 0,
    ymax = 80,
    xtick = data,
    xticklabels = {\textsc{Llama-2-7B}, \textsc{Llama-2-13B}, \textsc{Llama-2-70B}, \textsc{Falcon-180B}, \textsc{GPT-3.5}, \textsc{GPT-4}},
    ytick distance=10,
    ymajorgrids = true,
    enlarge x limits = {abs = .4},
    enlarge y limits = {abs = .02, upper},
    legend cell align=left,
    legend style={legend columns=5},
    height=5.9cm,
    width=16.8cm,
]
\addplot+ [
    error bars/.cd,
    y dir=both, y explicit,
] coordinates {
(0,39.1) += (0,4.2) -= (0,4.1)
(1,43.7) += (0,4.3) -= (0,4.3)
(2,49.1) += (0,4) -= (0,4.3)
(3,39.6) += (0,4.1) -= (0,4)
(4,43.1) += (0,4.1) -= (0,4)   
(5,55.9) += (0,4.3) -= (0,4.2)
};
\addplot+ [
    error bars/.cd,
    y dir=both, y explicit,
] coordinates {
(0,32) += (0,4.1) -= (0,3.9)
(1,42.4) += (0,4.1) -= (0,4.1)
(2,42.8) += (0,4.1) -= (0,4.1)
(3,38.1) += (0,3.8) -= (0,4.3)
(4,29.3) += (0,3.8) -= (0,3.7)
(5,36.5) += (0,4.1) -= (0,3.9) 
};
\addplot+ [
    error bars/.cd,
    y dir=both, y explicit,
] coordinates {
(0,43.7) += (0,4.3) -= (0,4.3)
(1,39.6) += (0,4.3) -= (0,4.3)
(2,48.3) += (0,4.3) -= (0,4.2)
(3,34.3) += (0,4) -= (0,3.9)
(4,32.2) += (0,3.9) -= (0,3.9)
(5,30) += (0,4.1) -= (0,3.7)
};
\addplot+ [
    error bars/.cd,
    y dir=both, y explicit,
] coordinates {
(0,29.4) += (0,3.9) -= (0,3.7)
(1,39.8) += (0,4.3) -= (0,4.1)
(2,45.7) += (0,4.3) -= (0,4.2)
(3,38.1) += (0,4.1) -= (0,4)
(4,45.7) += (0,4.5) -= (0,4.2)
(5,51.5) += (0,4.4) -= (0,4.1)
};
\addplot+ [
    fill opacity=0.6,
    error bars/.cd,
    y dir=both, y explicit,
] coordinates {
(0,16.1) += (0,3.2) -= (0,3)
(1,20.7) += (0,3.7) -= (0,3.3)
(2,16.5) += (0,3.3) -= (0,3)
(3,12.8) += (0,2.8) -= (0,2.6)
(4,12) += (0,3) -= (0,2.4)
(5,15.9) += (0,3.2) -= (0,2.9)
};

\legend {Factual $G_f$, Empty-Def $G_{f,\varepsilon}$, Counterfactual $G_c$, $G_{OOD}$, Empty-Def $G_{OOD,\varepsilon}$};

\coordinate (A) at (axis cs:0,16.667);
\coordinate (O1) at (rel axis cs:0,0);
\coordinate (O2) at (rel axis cs:5,0);
\draw [black,sharp plot,dashed] (A -| O1) -- (A -| O2);

\end{axis}

\end{tikzpicture}

%% file: images/scientific_concept_permutations.tikz
\begin{tikzpicture}
\begin{axis}[
    xmin=-0.3, xmax=5.3, ymin=8, ymax=46,
    xlabel={Number of counterfactual scientific concepts},
    ylabel={Classification accuracy},
    ylabel near ticks,
    ytick={0,10,20,25,30,35,40,45,50},
    ymajorgrids = true,
    legend pos=south west,
    width=8.1cm,
    height=6.2cm,
    legend cell align=left,
    legend style={legend columns=1, draw=none, fill=none, at={(0.05,0.06)},anchor=south west},
]
\addplot+ [
    sharp plot,
    error bars/.cd,
        y dir=both, y explicit,
] coordinates {
    (0,42.02) +-= (0,1.3448)
    (2,38.7) +-= (0,3.7574)
    (3,37.54) +-= (0,1.9115)
    (4,32.56) +-= (0,3.4173)
    (5,33.06) +-= (0,4.2615) 
};

\addplot+ [
    mark=diamond*,
    mark size=3.0pt,
    sharp plot,
    error bars/.cd,
        y dir=both, y explicit,
] coordinates {
    (0,27.36) +-= (0,4.4570)
    (2,28.16) +-= (0,3.5998)
    (3,26.68) +-= (0,5.9136)
    (4,26.14) +-= (0,2.4332)
    (5,26.66) +-= (0,4.4612) 
};
\legend {\textsc{GPT-3.5} ($r = -0.73$), \textsc{Llama-7B-chat} ($r = -0.10$)};
\end{axis}
\end{tikzpicture}

%% file: images/financial_concept_permutations.tikz
\begin{tikzpicture}
\begin{axis}[
    xmin=-0.3, xmax=6.3, ymin=28, ymax=53,
    xlabel={Number of counterfactual financial concepts},
    ylabel={Classification accuracy},
    ylabel near ticks,
    ytick distance=5,
    ymajorgrids = true,
    legend pos=south west,
    width=8.1cm,
    height=6.2cm,
    legend cell align=left,
    legend style={legend columns=1, draw=none, fill=none, at={(0.05,0.06)},anchor=south west},
]
\addplot+ [
    sharp plot,
    error bars/.cd,
        y dir=both, y explicit,
] coordinates {
    (0,46.2407) +-= (0,1.8906)
    (2,44.65) +-= (0,3.7574)
    (3,40.70) +-= (0,1.9115)
    (4,40.00) +-= (0,3.4173)
    (5,38.65) +-= (0,4.2615)
    (6,39.67) +-= (0,4.2615)
};

\addplot+ [
    mark=diamond*,
    mark size=3.0pt,
    sharp plot,
    error bars/.cd,
        y dir=both, y explicit,
] coordinates {
    (0,46.0926) +-= (0,3.2739)
    (2,46.6296) +-= (0,4.3647)
    (3,43.6667) +-= (0,3.1678)
    (4,42.8333) +-= (0,3.1789)
    (5,44.6296) +-= (0,2.8234)
    (6,42.8333) +-= (0,3.3277)
};

\legend {\textsc{GPT-3.5} ($r = -0.58$), \textsc{Llama-7B-chat} ($r = -0.32$)};
\end{axis}
\end{tikzpicture}

%% file: images/financial_human_guidelines.tikz
\begin{tikzpicture}
 
\begin{axis} [ybar = .05cm,
    bar width = 10.4pt,
    xmin = 0,
    xmax = 2,
    ymin = 0,
    ymax = 100,
    xtick = data,
    xticklabels = {\textsc{Llama-2-7B}, \textsc{GPT-3.5}, \textsc{GPT-4}},
    ytick distance=10,
    ymajorgrids = true,
    enlarge x limits = {abs = .5},
    enlarge y limits = {abs = .02, upper},
    legend cell align=left,
    legend style={legend columns=2, font=\small},
    height=5.9cm,
    width=8.45cm,
]
\addplot+ [
    error bars/.cd,
    y dir=both, y explicit,
] coordinates {
(0,48.9) += (0,4.2) -= (0,4.1)
(1,45.6) += (0,4.2) -= (0,4.3)   
(2,51.1) += (0,4.3) -= (0,4.1)
};
\addplot+ [
    error bars/.cd,
    y dir=both, y explicit,
] coordinates {
(0,39.1) += (0,4.0) -= (0,3.9)
(1,42.0) += (0,4.3) -= (0,4)
(2,47.2) += (0,4.3) -= (0,4.1) 
};
\addplot+ [
    error bars/.cd,
    y dir=both, y explicit,
] coordinates {
(0,39.3) += (0,4.0) -= (0,4.1)
(1,44.1) += (0,4.2) -= (0,4.3)
(2,48.1) += (0,4.3) -= (0,4)
};
\addplot+ [
    error bars/.cd,
    y dir=both, y explicit,
] coordinates {
(0,26.5) += (0,3.9) -= (0,3.5)
(1,40.6) += (0,4.2) -= (0,4.1)
(2,57.4) += (0,4.1) -= (0,4.3)
};
\addplot+ [
    fill opacity=0.6,
    error bars/.cd,
    y dir=both, y explicit,
] coordinates {
(0,29.8) += (0,4.1) -= (0,3.7)
(1,28.5) += (0,3.9) -= (0,3.7)
(2,32.4) += (0,4.1) -= (0,3.9)
};

\legend {Factual $G_f$, Empty-Def $G_{f,\varepsilon}$, Counterfactual $G_c$, $G_{OOD}$, Empty-Def $G_{OOD,\varepsilon}$};

\coordinate (A) at (axis cs:0,16.667);
\coordinate (O1) at (rel axis cs:0,0);
\coordinate (O2) at (rel axis cs:5,0);
\draw [black,sharp plot,dashed] (A -| O1) -- (A -| O2);

\end{axis}

\end{tikzpicture}